\documentclass{llncs}

\usepackage{amssymb}
\usepackage{amsmath}
\usepackage{subfigure}
\usepackage{graphicx}
\usepackage{multirow}

\begin{document}

\title{Topological descriptors for 3D surface analysis}
\titlerunning{Topological descriptors for 3D surface analysis}
\author{Matthias Zeppelzauer\inst{1} \and Bartosz Zieli\'{n}ski\inst{2} \and Mateusz Juda\inst{2} \and Markus Seidl\inst{1}}
\authorrunning{Matthias Zeppelzauer et al.}
\tocauthor{Matthias Zeppelzauer, Bartosz Zieli\'{n}ski, Mateusz Juda, and Markus Seidl}
\institute{Media Computing Group, Institute of Creative Media Technologies,\\
St. Poelten University of Applied Sciences,\\
Matthias-Corvinus Strasse 15, 3100 St. Poelten, Austra,\\
\email{\{m.zeppelzauer|markus.seidl\}@fhstp.ac.at},
\and
The Institute of Computer Science and Computer Mathematics,\\ Faculty of Mathematics and Computer Science, Jagiellonian University,\\
ul. {\L{}}ojasiewicza 6, 30-348 Krak{\'o}w, Poland,\\
\email{\{bartosz.zielinski|mateusz.juda\}@uj.edu.pl}}

\maketitle

\begin{abstract}
We investigate topological descriptors for 3D surface analysis, i.e. the classification of surfaces according to their geometric fine structure. On a dataset of high-resolution 3D surface reconstructions  we compute persistence diagrams for a 2D cubical filtration. In the next step we investigate different topological descriptors and measure their ability to discriminate structurally different 3D surface patches. We evaluate their sensitivity to different parameters and compare the performance of the resulting topological descriptors to alternative (non-topological) descriptors. We present a comprehensive evaluation that shows that topological descriptors are (i) robust, (ii) yield state-of-the-art performance for the task of 3D surface analysis and (iii) improve classification performance when combined with non-topological descriptors.
\end{abstract}

\section{Introduction}

With the increasing availability of high-resolution 3D scans, topological surface description is becoming increasingly important. In recent years, methods for sparse and dense 3D scene reconstruction have progressed strongly due to availability of inexpensive, off-the-shelf hardware (e.g. Microsoft Kinect) and the development of robust reconstruction algorithms (e.g. structure from motion techniques, SfM)~\cite{Crandall2011,Wu2013}. Since 3D scanning has become an affordable process the amount of available 3D data has increased significantly. At the same time, the reconstruction accuracy has increased strongly, which enables 3D reconstructions with sub-millimeter resolution~\cite{Wohlfeil2013}. The high resolution enables the accurate description of a 3D surface's geometric micro-structure, which opens up new opportunities for search and retrieval in 3D scenes, such as the recognition of objects by their specific surface properties as well as the distinction of different types of materials for improved scene understanding.

In this paper, we investigate the problem of describing and classifying 3D surfaces according to their geometric micro-structure. 
Two different types of approaches exist for this problem. Firstly, the dense processing of the surface in 3D space and secondly, the processing of the surface geometry in image-space based on depth maps derived from the surface. 

For the representation of surface geometry in 3D, descriptors are required that capture the local geometry around a given point or mesh vertex. Different types of local 3D descriptors have been developed recently that are suitable for the description of the local geometry around a 3D point, such as spin images~\cite{johnson1999using}, 3D shape context \cite{belongie2002shape}, and persistent point feature histograms \cite{rusu_persistent_2008}.

The dense extraction of surface geometry by local 3D descriptors, however,, however, becomes a computationally demanding task when several millions of points need to be processed. A computationally more efficient approach is the analysis of 3D surfaces in image space. In such approaches a 3D surface is first mapped to a depth map which represents a height field of the surface. This processing step maps the 3D surface analysis problem to a 2D texture analysis task which can be approached by analyzing the surface by texture descriptors, such as HOG, GLCM, and Wavelet-based features \cite{othmani2013single,zeppelzauerDH2015,zeppelzauer2015efficient}.

The presented approach falls into the category of image-space approaches. We first map the surface to image-space by a depth projection. Next, we divide the resulting depth map into patches and describe them with traditional non-topological as well as with topological surface descriptors. For the classification of surface patches we use random undersampling boosting (RUSBoost) \cite{seiffert2010rusboost} due to its high accuracy for imbalanced class distributions \cite{lopez2013insight}.

\section{Topological approach}
\label{sec:approach}

Mathematical standards topology, with its 120 years of history, is a relatively young discipline. It grew out of H. Poincare’s seminal work on the stability of the solar system as a qualitative tool to study the dynamics of differential equations without explicit formulas for solutions \cite{Poinc1890,Poinc1899,Poinc1895}. Due to the lack of useful analytic methods, topology soon became a purely theoretical discipline. However, in recent several years we observe an rapid development of topological data analysis tools, which open new applications for topology.

Topological spaces appearing in data analysis are typically constructed from small pieces or cells.
A natural tool in the study of multidimensional images with topological methods are hypercubes (points, edges, squares, cubes etc.), e.g. a pixel in a $2$ dimensional image is equivalent to a square, a voxel in a $3$ dimensional volume is equivalent to a cube. Hypercubes are building blocks for structures called cubical complexes. Such representations give topology a combinatorial flavour and make it a natural tool in the study of multi-dimensional data sets.

Intuitively, the rank of the $n$th homology group, the so called $n$th Betti
number denoted $\beta_n$, counts the number of $n$-dimensional holes in the topological space.
In particular, $\beta_0$ counts the number of connected components.
As an example consider the image of the digit “8”. In this image
there is one connected component and two holes, hence $\beta_0=1$ and $\beta_1 =2$.
For a hollow sphere we have $\beta_0 =1$, $\beta_1 =0$, $\beta_2 =1$.
For a tube in a tire we have $\beta_0 =1$, $\beta_1 =2$, $\beta_2 =1$.

Betti numbers do not differentiate between small and large holes.
In consequence, the holes resulting from the noise in the data cannot be
distinguished from the holes indicative for the nature of the data.
For instance, in a noisy image of the digit “8” one can get easily $\beta_0 > 1$.
A remedy for this drawback is persistent homology,
a tool invented at the beginning of the 20th century \cite{EdLeZo2002}.
Persistent homology studies how the Betti numbers change when
the topological space is gradually built by adding cubes in some prescribed order.

If $X$ is a cubical complex, one can add cubes step by step.
Typically, the construction goes through different scales, starting from the smallest pieces.
However, in general an arbitrary function $f:X\to\mathbb{R}$, called the Morse function or measurement function,
may be used to control the order in which the complex is built, starting from low values of $f$ and increasing subsequently.
This way we obtain a sequence of topological spaces, called a filtration,
\[
    \emptyset=X_{r_0}\subset X_{r_1}\subset X_{r_2}\subset\cdots\subset X_{r_n}=X,
\]
where $X_r:=f^{-1}((-\infty,r])$ and $r_i$ is a growing sequence of values of $f$
at which the complex changes.
As the space is gradually constructed, holes are born, persist for some time
and eventually may die. The length of the associated birth-death intervals (persistence intervals) indicates if the holes are relevant or merely noise.
The lifetime of holes is usually visualized by the so called persistence diagram (PD). Persistence diagrams constitute the main tool of topological data analysis. They visualize geometrical properties of a multidimensional object $X$ in a simple two dimensional diagram.

\begin{figure}%
\centering
	\subfigure[]{\label{patch:org}
	\includegraphics[width=0.31\linewidth]{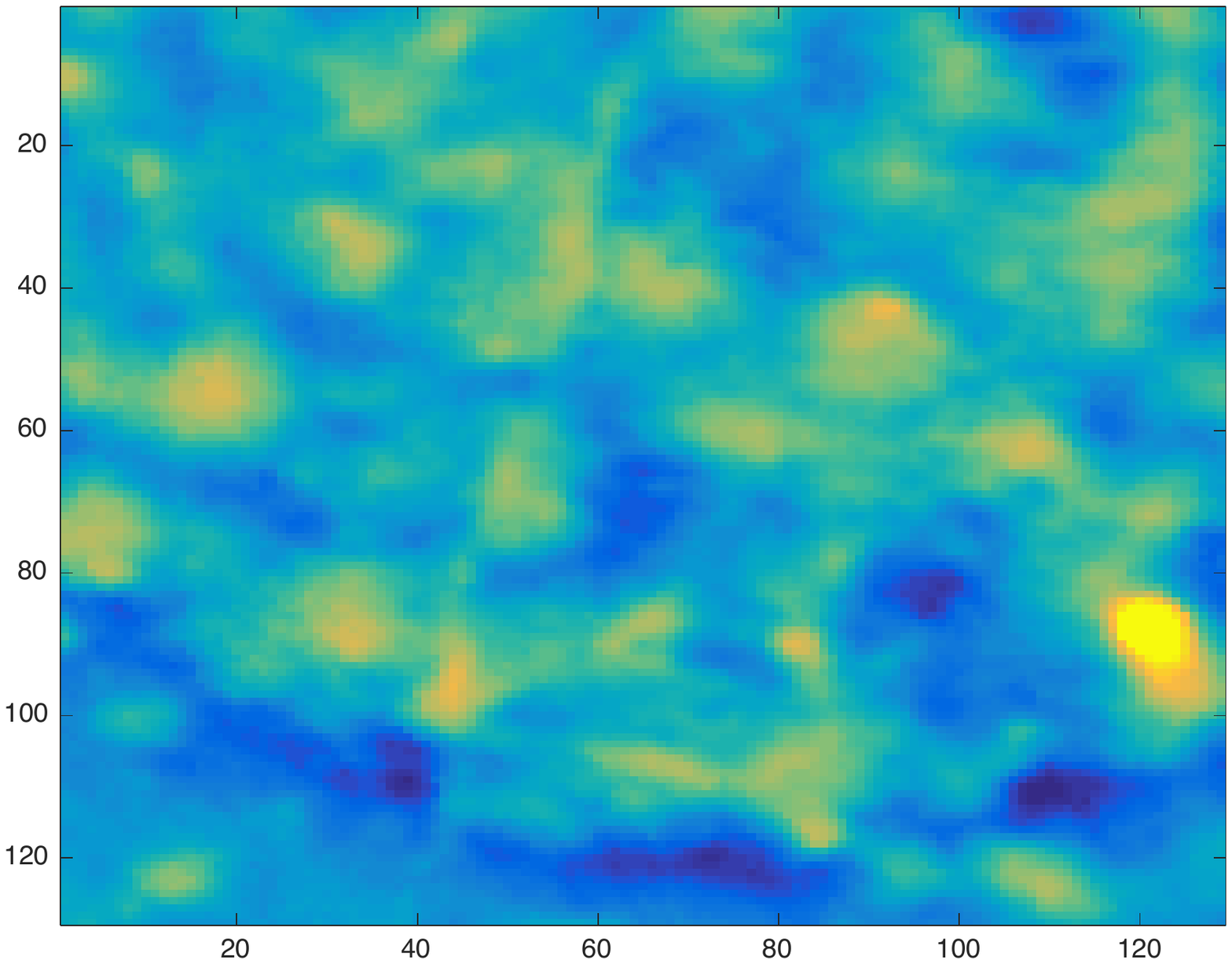}}
	\subfigure[]{\label{patch:pd}
	\includegraphics[width=0.31\linewidth]{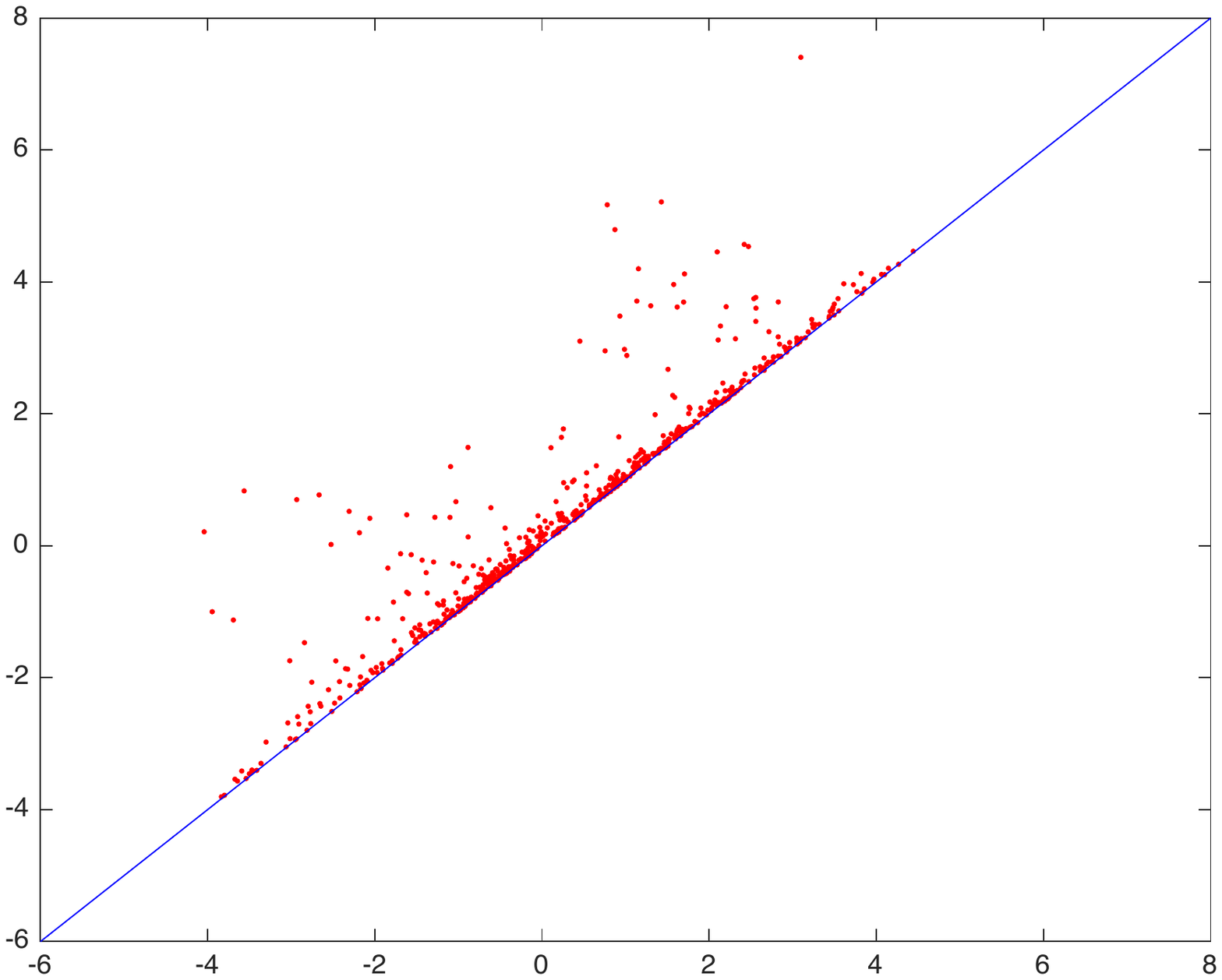}}
	\subfigure[]{\label{patch:pi}
	\includegraphics[width=0.31\linewidth]{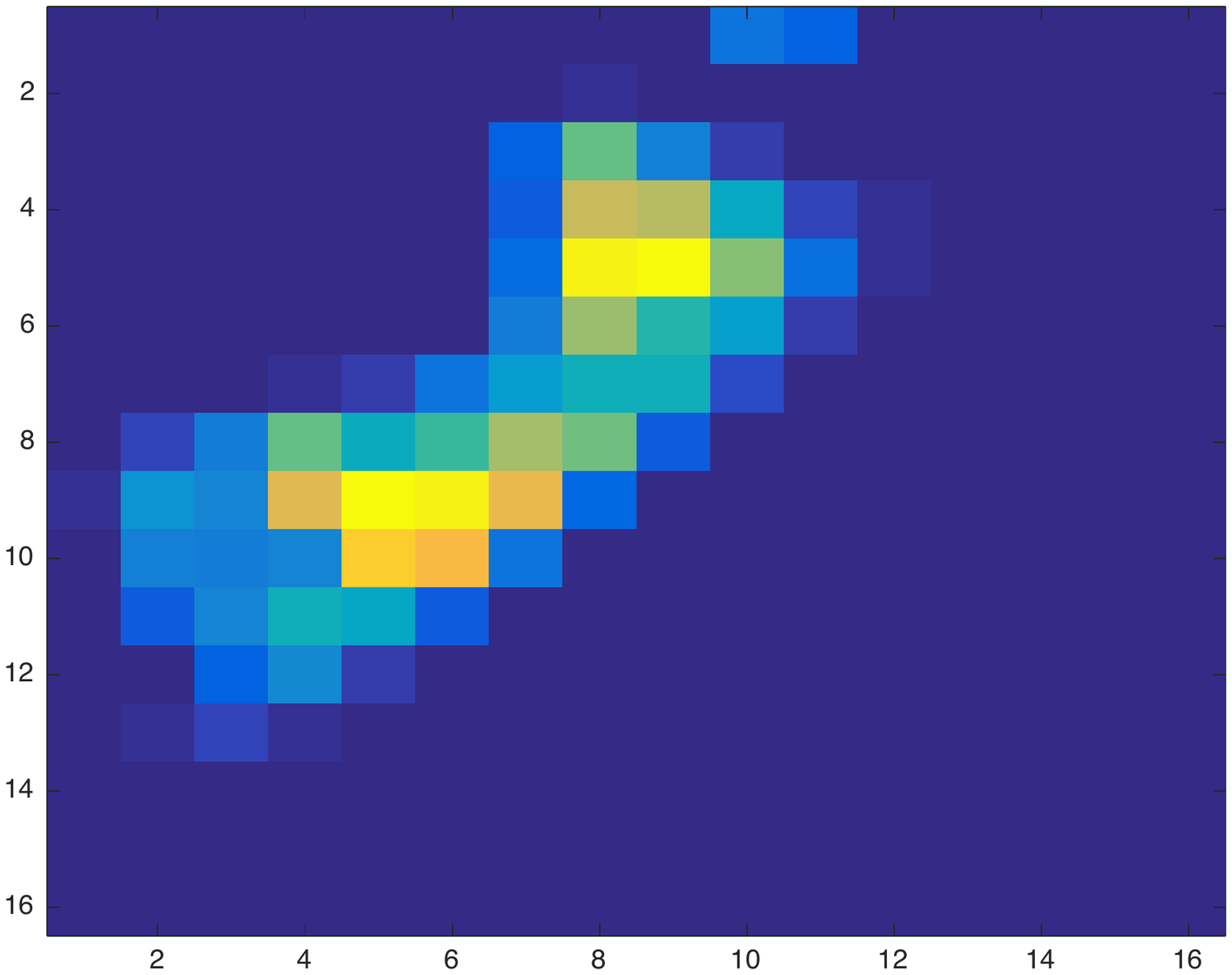}}
	\caption{Example patch: (a) the original 3D surface as a 2D depth map; (b) the corresponding persistent diagram; (c) and the persistent image with $\sigma=0.001$ and resolution $16\times16$.}
	\label{fig:patch}
\end{figure}

Figure~\ref{fig:patch}(a) shows a 3D surface as a 2D depth map, where colors corresponds to depth (blue refers to low depth, yellow to high depth).
In this case pixels are represented as 2-dimensional cells of a cubical complex. For the complex we can obtain a filtration $X_r$ using a measuring function which has a value for a 2-dimensional cube equal to height (pixel color). For a lower dimensional cell (a vertex or an edge) we can set the function value as a maximum from the higher-dimensional neighborhoods of the cell. Figure~\ref{fig:patch}(b) shows the persistence diagram for $X_r$.

There is still no concrete answer on how and when the tools of computational topology and machine learning should be used together.
A first attempt is to provide a descriptor of a topological space filtration based on elementary statistics of persistence intervals (or equivalently on persistence diagrams). Let
\[
   I := \{[b_1,e_1], [b_2,e_2], \ldots, [b_n,e_n] \}
\]
be a set of persistence intervals. Let $D := \{ d_i := (e_i - b_i)\}_{i=1}^n$ be a set of the interval lengths. 
We build an aggregated descriptor of $D$, denoted by PD\_AGG, using following measures:
number of elements, minimum, maximum, mean, standard deviation, variance, $1$-quartile, median, $3$-quartile, and norms $\sum\sqrt{d_i}$, $\sum d_i$, and $\sum(d_i)^2$.

Except the PD\_AGG descriptor described above, which can be used with standard classification methods, there are also attempts to use PD directly with appropriately modified classifiers. Reininghaus et al. \cite{reininghaus2014stable} proposed a multiscale kernel for PDs, which can be used with a support vector machine (SVM). While this kernel is well-defined in theory, in practice it becomes highly inefficient for a large number of training vectors (as the entire kernel matrix must be computed explicitly). As an alternative, Chepushtanova et al. \cite{chepushtanova2015persistence} introduced a novel representation of a PD, called a persistence image (PI), which is faster and can be use with a broader range of machine learning (ML) techniques.

A PI is derived from mapping a PD to an integrable function $G_p: \mathbb{R}^2 \rightarrow \mathbb{R}$, which is a sum of Gaussian functions centered at each point of the PD. Taking a discretization of a subdomain of $G_p$ defines a grid. An image can be created by computing the integral of $G_p$ on each grid box, thus defining a matrix of pixel values. Formally, the value of each pixel $p$ within a PI is defined by the following equation:
\[
PI(p) = \iint\limits_{p} \sum_{[b_i, e_i] \in I}{g(b_i, e_i) \, \dfrac{1}{2 \pi \sigma_x \sigma_y} \, e^{-\frac{1}{2} \left( \frac{(x - b_i)}{\sigma_x^2} + \frac{(y - e_i)}{\sigma_y^2} \right) }} \,dy\,dx,
\]
where $g(b_i, e_i)$ is a weighting function, which depends on the distance from the diagonal (points close to the diagonal are usually considered as noise, therefore they should have low weights), $\sigma_x$ and $\sigma_y$ are the standard deviations of the Gaussians in $x$ and $y$ direction. The resulting image (see Figure \ref{patch:pi}) is vectorized to achieve a standardized vectorial representation which is compatible to a broad range of ML techniques.

The advantage of PIs compared to PDs descriptor is a high classification accuracy, however they are unstable according to \cite{chepushtanova2015persistence}. Moreover, they require numerous parameters like the PI resolution, the weighting function $g$, as well as $\sigma_x$ and $\sigma_y$.

\section{Experimental Setup}
\label{sec:setup}

\begin{figure}%
\centering
	\subfigure[]{\label{sfig:pc}
	\includegraphics[height=0.31\linewidth, angle=90]{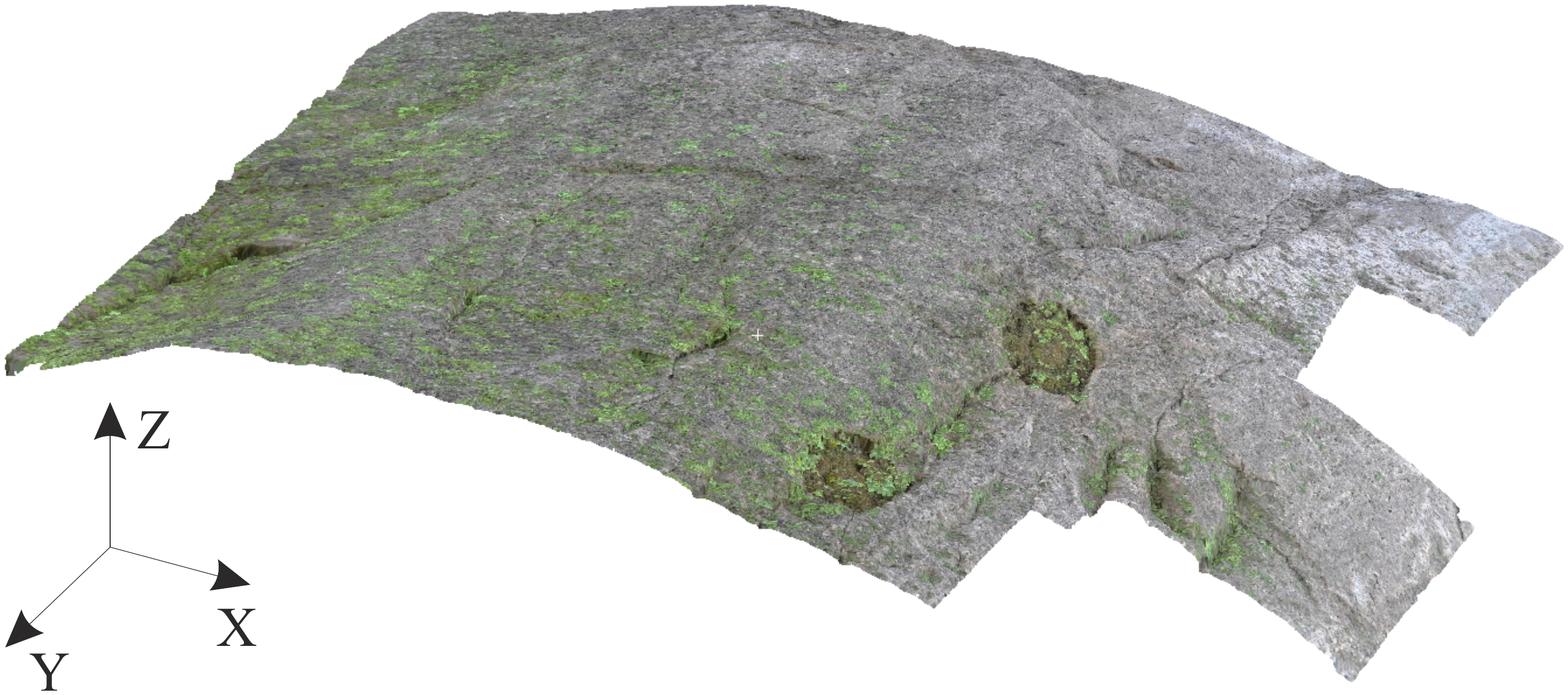}}
	\subfigure[]{\label{sfig:flatDM}
	\includegraphics[width=0.31\linewidth]{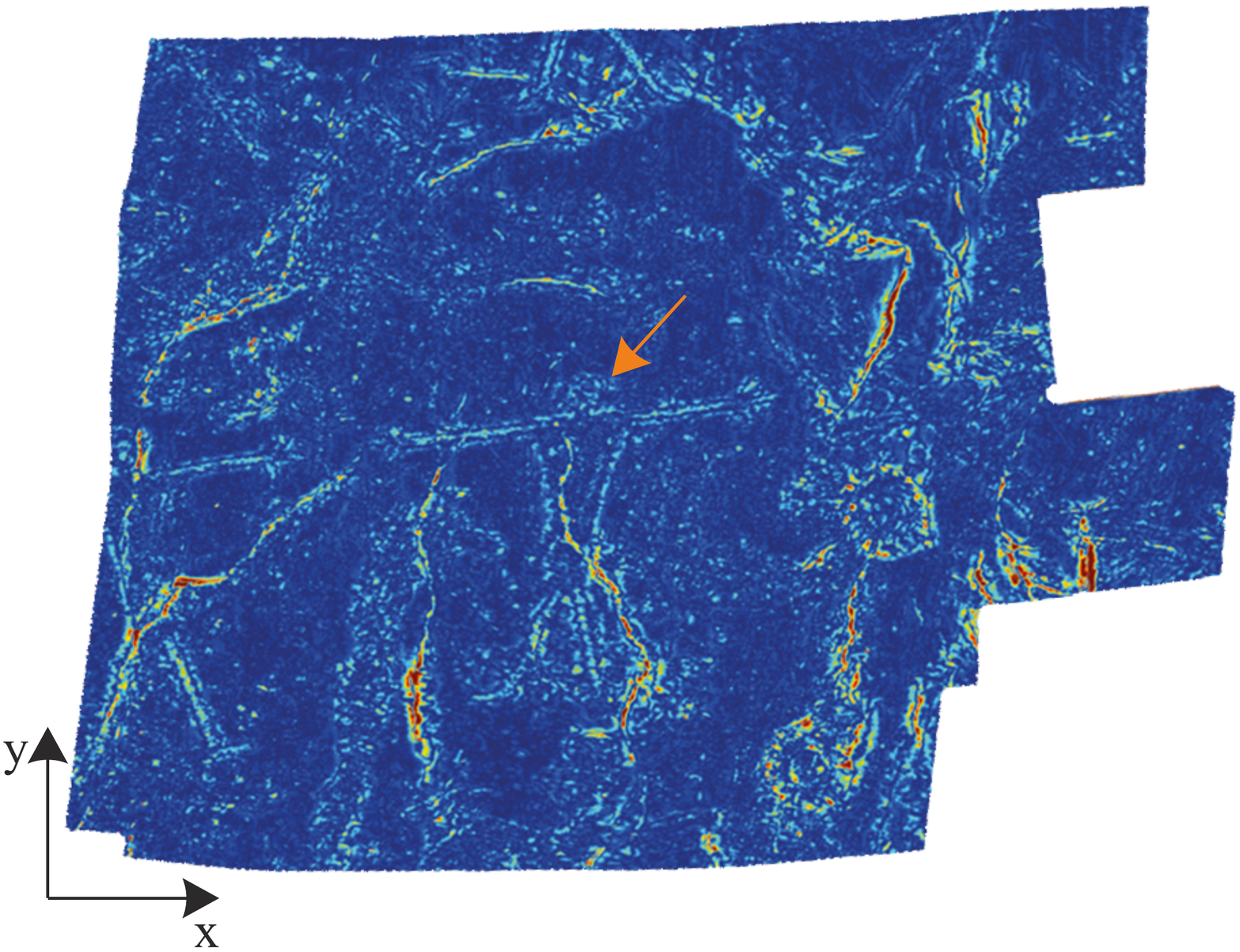}}
	\subfigure[]{\label{sfig:anno}
	\includegraphics[width=0.31\linewidth]{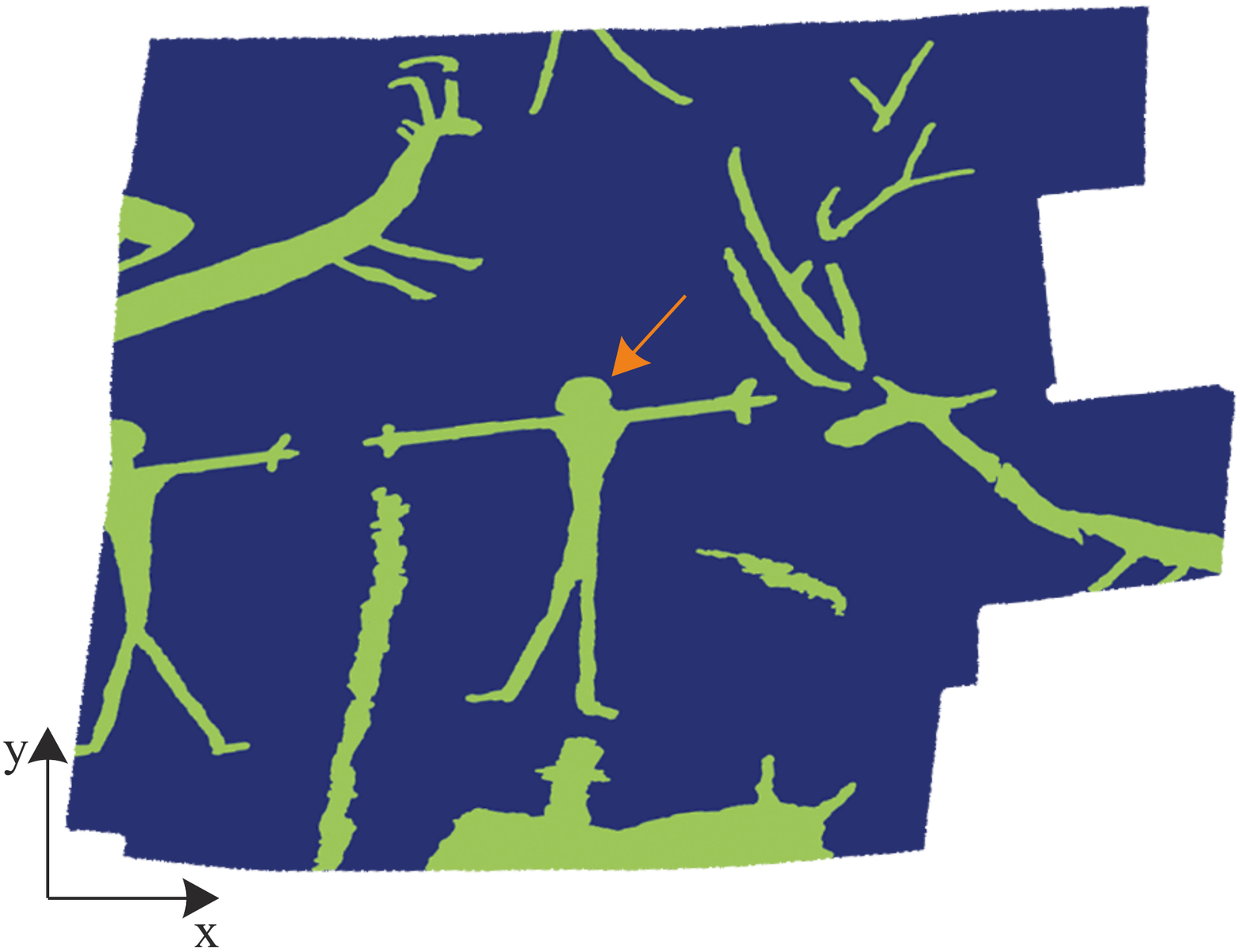}}
	\caption{Example data: (a) the 3D point cloud of the surface; (b) the depth projection of the surface with compensated global curvature; (c) ground truth labeling that specifies areas with different topography, such as the human-shaped figure in the center whose head is marked with an arrow.}
	\label{fig:depthMap}
\end{figure}

In our experiments we investigate the robustness and expressiveness of the topological descriptors presented in Section \ref{sec:approach} for 3D surface analysis and compare and combine them with traditional non-topological descriptors.
For our experiments, we employ a dataset of high-resolution 3D reconstructions from the archaeological domain with a resolution below 0.1mm \cite{zeppelzauer2015efficient}. The dimension of the scanned surfaces ranges from approx. $20 \times 30$ cm to $30 \times 50$ cm. The reconstructions represent natural rock surfaces that exhibit human-made engravings (so-called rock-art). The engravings represent symbols and figures (e.g. animals and humans) engraved by humans in ancient times. See Figure~\ref{fig:depthMap} for an example surface.
The engraved regions in the surface exhibit a different surface geometry than the surrounding natural rock surface.   In our experiments we aim at automatically separating the engraved areas from the natural rock surface. The corresponding ground truth is depicted in Figure \ref{sfig:anno}.

The employed dataset contains 4 surface reconstructions with a total number of 12.3 millions of points. For each surface a precise ground truth has been generated by domain experts that labels all engravings on the surface. The dataset contains two classes of surface topographies: class 1 represents engraved areas and class 2 represents the natural rock surface. Class priors are imbalanced. Class 1 represents 16.6\% of the data and is thus underrepresented.

For each scan we perform depth projection and preprocessing as described in \cite{zeppelzauer2015efficient}. The result is a depth map that reflects the geometric micro-structure of the surface, see Figure \ref{sfig:flatDM}. This representation is the input to feature extraction.

From the depth map we extract a number of non-topological image descriptors in a block-based manner that serve as a baseline in our experiments. The block size is $128 \times 128$ pixels (i.e. $10.8 \times 10.8$ mm) and the step size between two blocks is 16 pixels (1.35 mm). The baseline features include: MPEG-7 Edge Histogram (EH) \cite{mpeg7standard}, Dense SIFT (DSIFT) \cite{lowe2004distinctive}, Local Binary Patterns (LBP) \cite{ojala1996comparative}, Histogram of Oriented Gradients (HOG) \cite{dalal2005histograms}, Gray-Level Co-occurrence Matrix (GLCM) \cite{haralick1973textural}, Global Histogram Shape (GHS), Spatial Depth Distribution (SDD), as well as manually modified enhanced versions of GHS and SDD (short EGHS and ESDD) that apply additional enhancements to the depth map described in \cite{zeppelzauer2015efficient}.

Additionally to the baseline descriptors, we extract persistent homology descriptors in the same block-wise manner. For each patch, we compute a persistence diagram and derive the 12-dimensional aggregated descriptor (PD\_AGG) as described in Section \ref{sec:approach}. Additionally, we extract persistence images (PIs) for different resolutions (8, 16, 32, 64) and standard deviations (0.00025, 0.0005, 0.001, 0.002) with and without weighting (see Section \ref{sec:approach}).

Alternatively, we first extract Completed LBP (CLBP) features \cite{guo2010completed} from the depth map as proposed in \cite{li2014persistence} and \cite{reininghaus2014stable} and then extract PD\_AGG and PIs from the CLBP\_S and CLBP\_M maps.

After feature extraction the entire dataset is split into independent training and evaluation sets. The training set contains image patches from scans 1 and 2 from the dataset. Scans 3 and 4 make up the evaluation set. From the training set we randomly select 50\% of the blocks from class 1 (2962 blocks) and 30\% from class 2 (7592 blocks). On this subset of 9654 samples, we apply 5-fold cross-validation to estimate suitable classifier parameters. The best parameters are used to train the classifier on the entire training set. The trained classifier is finally applied to the independent evaluation set of 27192 patches. As our dataset has imbalanced classes, we employ RusBoost \cite{seiffert2010rusboost} in our experiments.

As a performance measure we employ the Dice Similarity Coefficient (DSC). DSC measures the mutual overlap between an automatic labeling $X$ of an image and a manual (ground truth) labeling $Y$:
\[
\text{DSC}(X,Y)=\frac{2 |X \cap Y|}{|X|+|Y|}.
\]
DSC is between $0$ and $1$ where $1$ means a perfect segmentation.

Each classification experiment is repeated 10 times with 10 different randomly selected subsets from the training set to reduce the dependency from the training data. From the 10 resulting DSC values we provide median and standard deviation as the final performance measures. 

Aside from quantitative evaluations we investigate the following questions:
\begin{itemize}
	\item Can persistent homology descriptors outperform descriptors like HOG, SIFT, and GLCM for surface classification?
    \item How does aggregation of the PD (PD\_AGG) influence performance compared to non-aggregated representations like PI?
    \item Is CLBP a suitable input representation for persistent homology descriptors?
    \item How sensitive is PI to its parameters (resolution, sigma, weighting)?
    \item Do persistent homology descriptors add beneficial or even necessary information to the baseline descriptors in our classification task? 
\end{itemize}

The experiment was implemented in Matlab. Most of the descriptors were extracted with VLFeat library \cite{vedaldi2010vlfeat}, except PD\_AGG and PI. We compute persistence intervals of the images using CAPD::RedHom library \cite{RedHom,juda2014capd} with the PHAT \cite{url:PHAT,PHAT} algorithm for persistence homology.

\section{Results}

We start our evaluation with the aggregated descriptor PD\_AGG. The descriptor applied to our surfaces yields a DSC of $0.6528\pm0.0118$ and represents a first baseline for further comparisons. Next, we apply PI with different resolutions, sigmas with and without weighting. Results are summarized in Table \ref{tab:PI}. All results for PI outperform that of PD\_AGG. We assume the reason is that PD\_AGG neglects the information about the points' localization, which is preserved in PI. The best result for PI is a DSC of $0.7335\pm0.0024$ without weighting. The difference between the best weighting and no weighting result is statistically significant\footnote{Statistical significance is computed with the Wilcox signed rank test, as most of the samples do not pass the Shapiro-Wilk normality test.} with $p-value=0.006$. This result is surprising as it is contrary to the results of \cite{chepushtanova2015persistence} where artificial datasets were used for evaluation. Results in Table \ref{tab:PI} further show that PI has low sensitivity to different resolutions and sigmas.

\begin{table}
  \caption{DSC for PI descriptors depending on the sigma of the Gaussian function ($\sigma$) and resolution ($res$). Bold represents the best results for PI with and without weighting.}
  \label{tab:PI}
  \begin{center}
  \resizebox{\textwidth}{!}{
  \begin{tabular}{l@{\hskip 0.1in}l@{\hskip 0.1in}c@{\hskip 0.1in}c@{\hskip 0.1in}c@{\hskip 0.1in}c}
  \hline\rule{0pt}{12pt}
   &  & $res=8\times8$ & $res=16\times16$ & $res=32\times32$ & $res=64\times64$ \\[2pt]
  \hline\rule{0pt}{12pt}
  \multirow{4}{*}{weighting} & $\sigma=0.00025$ & $0.714\pm0.005$ & $0.718\pm0.007$ & $0.715\pm0.007$ & $0.709\pm0.008$ \\
   & $\sigma=0.0005$ & $0.718\pm0.005$ & $0.715\pm0.005$ & $0.715\pm0.006$ & $0.714\pm0.004$ \\
   & $\sigma=0.001$ & $0.715\pm0.006$ & $0.716\pm0.005$ & $0.718\pm0.004$ & $\mathbf{0.718\pm0.005}$ \\
   & $\sigma=0.002$ & $0.706\pm0.003$ & $0.719\pm0.005$ & $0.715\pm0.004$ & $0.710\pm0.005$ \\[2pt]
  \hline\rule{0pt}{12pt}
  no wghting. & $\sigma=0.001$ & $0.724\pm0.004$ & $\mathbf{0.734\pm0.002}$ & $0.732\pm0.004$ & $0.733\pm0.004$ \\[2pt]
  \hline
  \end{tabular}
  }
  \end{center}
  
\end{table}

Next, we evaluate the performance of PD\_AGG and PI with CLBP as input representation, see Table \ref{tab:PDPILBP}. The best result for PD\_AGG ($0.6874\pm0.0030$) is obtained for the rotation invariant CLBP maps with radius 5 and number of samples 16. This improvement is statistically significant, with $p-value=0.002$ (compared to the PD\_AGG without CLBP). For PI we do not observe an improvement. This was confirmed by further experiments, where we combined PI obtained for the original depth map with PI on CLBP maps. The resulting DSC equals $0.7178\pm0.0034$. This  shows not only that CLBP brings no additional information for PI, but further  indicates that it can even be confusing for the classifier. The expressiveness of PI seems to be at a level where CLBP is not able to add additional information. Whereas PD\_AGG is less expressive and thus benefits from the additional processing.

\begin{table}
\caption{DSC for PD\_AGG and PI descriptors extracted from the CLBP\_S and CLBP\_M maps. We consider two encodings for CLBP: rotation invariant uniform (riu2) and rotation invariant (ri) and vary radius r and the number of samples n. Bold numbers represent the best results for PD\_AGG and PI.}
\label{tab:PDPILBP}
\begin{center}
\begin{tabular}{l@{\hskip 0.1in}l@{\hskip 0.1in}l@{\hskip 0.1in}c@{\hskip 0.1in}c}
\hline\rule{0pt}{12pt}
Descriptor & CLBP type &  & $n=8$ & $n=16$\\[2pt]
\hline\rule{0pt}{12pt}
\multirow{4}{*}{PD\_AGG}& \multirow{2}{*}{riu2} & $r=3$ & $0.613\pm0.009$ & $0.625\pm0.005$ \\
&  & $r=5$ & $0.654\pm0.003$ & $0.636\pm0.010$ \\[2pt]
\cline{2-5}\rule{0pt}{12pt}
& \multirow{2}{*}{ri} & $r=3$ & $0.632\pm0.009$ & $0.666\pm0.007$ \\
&  & $r=5$ & $0.681\pm0.004$ & $\mathbf{0.687\pm0.003}$ \\[2pt]
\hline\rule{0pt}{12pt}
\multirow{4}{*}{PI}& \multirow{2}{*}{riu2} & $r=3$ & $0.688\pm0.005$ & $0.702\pm0.004$ \\
&  & $r=5$ & $0.704\pm0.002$ & $0.717\pm0.003$ \\[2pt]
\cline{2-5}\rule{0pt}{12pt}
& \multirow{2}{*}{ri} & $r=3$ & $0.699\pm0.002$ & $0.699\pm0.002$ \\
&  & $r=5$ & $0.703\pm0.003$ & $\mathbf{0.703\pm0.003}$ \\[2pt]
\hline
\end{tabular}
\end{center}
\end{table}

\begin{figure} 
\centering
	\subfigure[]{\label{sfig:mapGini}
	\includegraphics[width=0.45\linewidth]{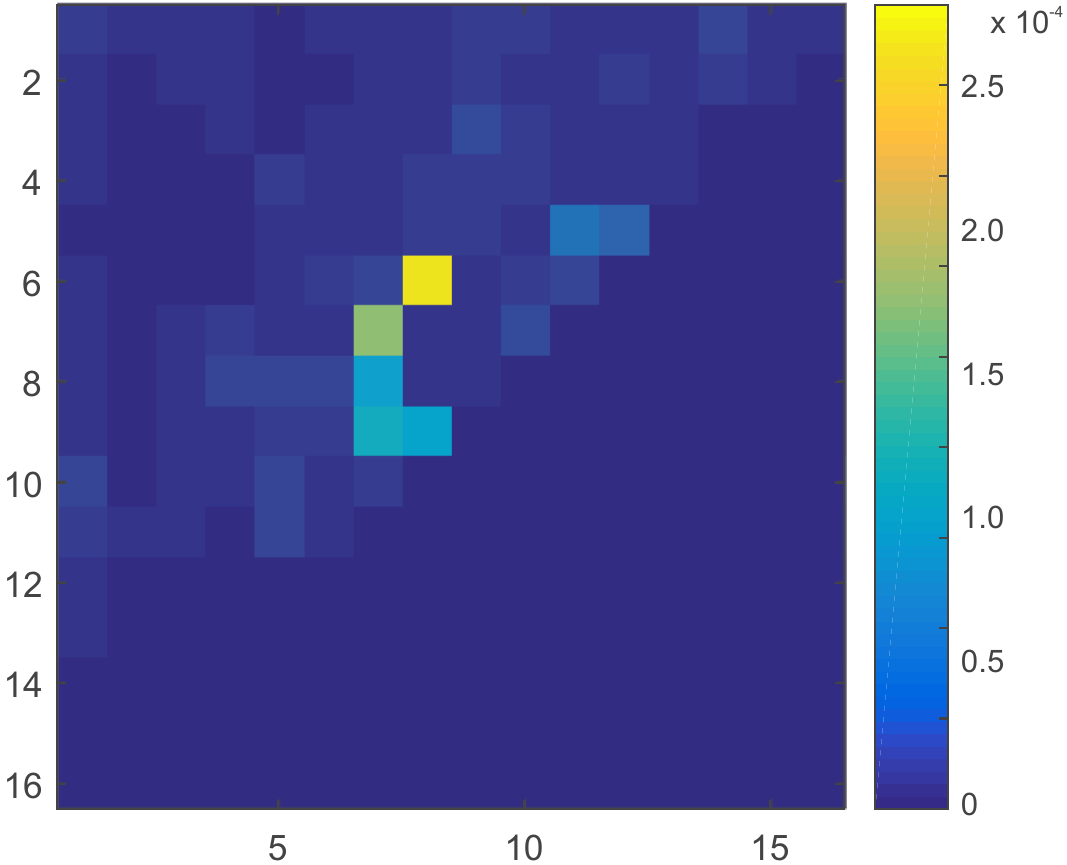}}
	\subfigure[]{\label{sfig:mapFisher}
	\includegraphics[width=0.425\linewidth]{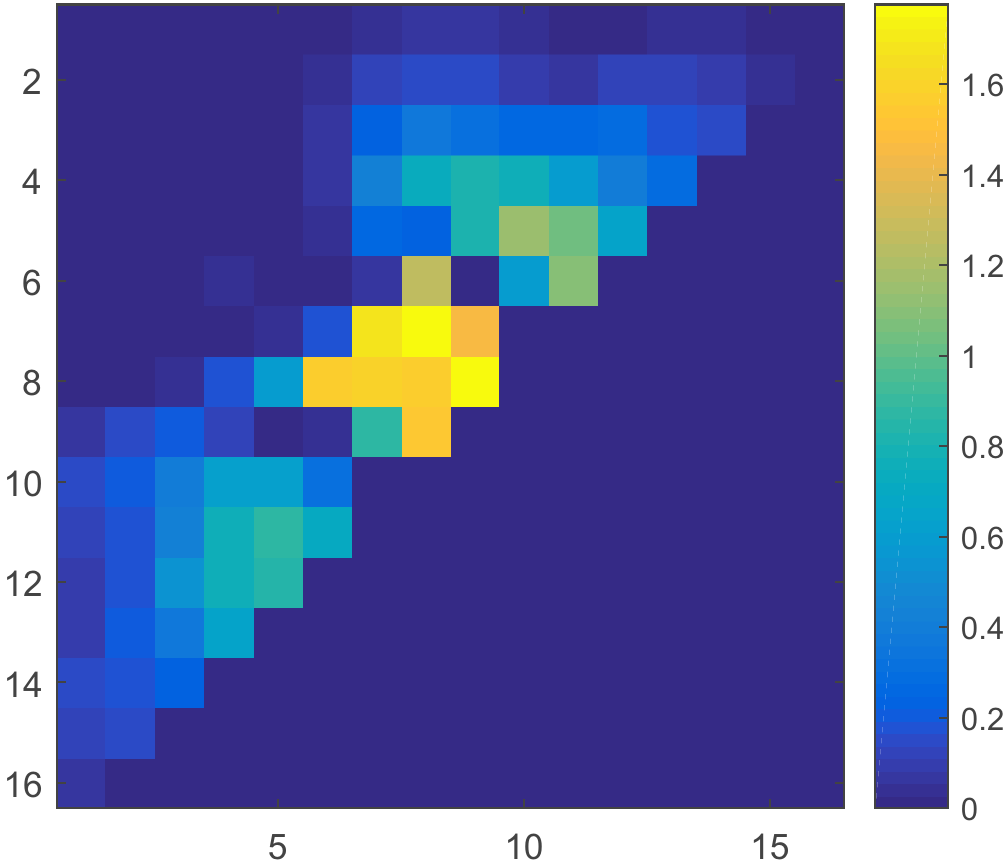}}
    \caption{Importance of the PI's pixels obtained with Gini importance measure and Fisher discriminant.}
	\label{fig:importanceMaps}
\end{figure}

As a next step we investigate which locations of PI are the most important ones for classification. For this purpose we computed Gini importance measure for each location of the PI, see Fig. \ref{sfig:mapGini}. The most important pixels are located in the middle of the PI. It is worth noting that there are only few very important pixels, while the others are more than 10 time less important. Moreover, there are few important pixels near to the center of the diagonal. To get a more complete picture, we compute the Fisher discriminant for each location of the PI, see Fig. \ref{sfig:mapFisher}. The result is to a large degree consistent with the Gini measure and confirms our observation.

Finally, we investigate the performance of topological vs. non-topological descriptors and their combinations. The DSC for baseline descriptors and for their combination with PD\_AGG and PI are presented in Table \ref{tab:BPDPI}. Our experiments show that both topological descriptors contribute additional valuable information to the baseline descriptors and improve the classification accuracy. All combinations with PD\_AGG are significantly better than the baseline itself. Moreover, PI works significantly better than PD\_AGG with all of the baseline descriptors (except for GHS, GHS+SDD, EGHS+ESDD where the improvement is not significant).

\begin{table}
\caption{DSC for baseline descriptors (B) and their combination with PD\_AGG and PI descriptors (B + PD\_AGG and B + PI, respectively). Asterisks ($^*$) correspond to $p-values < 0.01$ when comparing B to B + PD\_AGG and B + PD\_AGG to B + PI.}
\label{tab:BPDPI}
\begin{center}
\begin{tabular}{l@{\hskip 0.1in}c@{\hskip 0.1in}c@{\hskip 0.1in}c}
\hline\rule{0pt}{12pt}
Descriptor & Baseline (B) & B + PD\_AGG & B + PI\\[2pt]
\hline\rule{0pt}{12pt}
EH & $0.641\pm0.007$ & $0.669\pm0.015^*$ & $0.696\pm0.015^*$ \\
\hspace{0.3em}LBP & $0.452\pm0.020$ & $0.531\pm0.023^*$ & $0.587\pm0.027^*$ \\
\hspace{0.3em}DSIFT & $0.486\pm0.003$ & $0.739\pm0.004^*$ & $0.764\pm0.004^*$ \\
\hspace{0.3em}HOG & $0.503\pm0.008$ & $0.712\pm0.007^*$ & $0.732\pm0.003^*$ \\
\hspace{0.3em}GLCM & $0.645\pm0.003$ & $0.706\pm0.002^*$ & $0.732\pm0.002^*$ \\
\hspace{0.3em}GHS & $0.301\pm0.048$ & $0.470\pm0.038^*$ & $0.476\pm0.066$ \\
\hspace{0.3em}SDD & $0.692\pm0.003$ & $0.735\pm0.003^*$ & $0.767\pm0.004^*$ \\
\hspace{0.3em}GHS+SDD & $0.399\pm0.028$ & $0.426\pm0.027^*$ & $0.454\pm0.029$ \\
\hspace{0.3em}EGHS & $0.650\pm0.008$ & $0.683\pm0.003^*$ & $0.690\pm0.004^*$ \\
\hspace{0.3em}ESDD & $\mathbf{0.743\pm0.002}$ & $\mathbf{0.763\pm0.002^*}$ & $\mathbf{0.790\pm0.002^*}$ \\
\hspace{0.3em}EGHS+ESDD & $0.728\pm0.005$ & $0.740\pm0.003^*$ & $0.743\pm0.005$ \\[2pt]
\hline
\end{tabular}
\end{center}
\end{table}

\section{Conclusion}
We have presented an investigation of topological descriptors for 3D surface analysis. Our major conclusions are: (i) the aggregation of persistence diagrams removes important information which can be retained by using PI descriptors, (ii) PIs are expressive and robust descriptors that are well-suited to include topological information into ML pipelines, and (iii) topological descriptors are complementary to traditional image descriptors and represent necessary information to obtain peak performance in 3D surface classification. Furthermore, we observed that short intervals in the PD contribute more to classification accuracy than expected. This will be subject to future research.

\section{Acknowledgements}
Parts of the work for this paper has been carried out in the project 3D-Pitoti which is funded from the European Community's Seventh Framework Programme (FP7/2007-2013) under grant agreement no 600545; 2013-2016.

\bibliographystyle{splncs03}
\bibliography{mbm_ctic}

\end{document}